\documentclass{IEEEtran}
\usepackage{cite}
\usepackage[utf8]{inputenc}
\usepackage{amsmath,amssymb,amsfonts}
\usepackage{amsthm}
\usepackage{multirow, array}
\usepackage{cite}
\usepackage[T1]{fontenc}
\usepackage{breqn}
\usepackage{cite}
\usepackage{algorithmic}
\usepackage{graphicx}
\usepackage{textcomp}
\usepackage{xcolor}
\usepackage{float}
\usepackage{subcaption}
\usepackage{multirow}
\usepackage{booktabs}
\usepackage{csquotes}
\usepackage{amsmath}
\newtheorem{definition}{\textbf{Definition}}

\newtheorem{principle}{Principle}

\title{Exploring Core and Periphery Precepts in Biological and Artificial Intelligence: An Outcome-Based Perspective} 

 \author{Niloofar Shadab$\IEEEauthorrefmark{1}\IEEEauthorrefmark{2}$,
         Tyler Cody$\IEEEauthorrefmark{2}\IEEEauthorrefmark{3}$,
         Alejandro Salado$\IEEEauthorrefmark{4}$, Taylan G. Topcu$\IEEEauthorrefmark{2}$, Mohammad Shadab$\IEEEauthorrefmark{5}$,
         Peter Beling$\IEEEauthorrefmark{2}\IEEEauthorrefmark{3}$\\
         \small $\IEEEauthorrefmark{2}$Grado Department of Industrial and Systems Engineering, Virginia Tech \\
         \small $\IEEEauthorrefmark{3}$Responsible General Intelligence Lab, Virginia Tech \\
         \small $\IEEEauthorrefmark{4}$Department of Systems and Industrial Engineering, The University of Arizona \\
         \small $\IEEEauthorrefmark{5}$School of Computing, Binghamton University\\
         \small $\IEEEauthorrefmark{1}$Corresponding author: Niloofar Shadab; nshadab@vt.edu \\
}

\begin{document}

\maketitle

\begin{abstract}
Engineering methodologies predominantly revolve around established principles of decomposition and recomposition. These principles involve partitioning inputs and outputs at the component level, ensuring that the properties of individual components are preserved upon composition. However, this view does not transfer well to intelligent systems, particularly when addressing the scaling of intelligence as a system property. Our prior research contends that the engineering of general intelligence necessitates a fresh set of overarching systems principles. As a result, we introduced the "core and periphery" principles, a novel conceptual framework rooted in abstract systems theory and the Law of Requisite Variety. In this paper, we assert that these abstract concepts hold practical significance. Through empirical evidence, we illustrate their applicability to both biological and artificial intelligence systems, bridging abstract theory with real-world implementations. Then, we expand on our previous theoretical framework by mathematically defining core-dominant vs periphery-dominant systems.
\end{abstract}


\section{Introduction}

Engineering principles rely on decomposition mechanisms to "divide and conquer" complex problems by breaking them into smaller, manageable components, which collectively exhibit system-level properties when integrated \cite{simon2012architecture, parnas1972criteria}. Traditional decomposition methods effectively address system properties\cite{sage1998systems, eppinger_planning_1997, baldwin_design_2000}; however, these approaches are fundamentally rooted in the open systems paradigm \cite{wymore2018model}. This paradigm assumes that systems and their constituent elements operate based on predefined input-output relationships coordinated through well-defined interfaces. Nevertheless, complex systems are only partially decomposable \cite{simon2019sciences, topcu2022dark}, and the emergence of intelligent systems challenges the broader efficacy and generalizability of these traditional concepts \cite{shadab2024closed}.

In our previous work, we posited that intelligence is an attribute intrinsic to the interaction between the system and its context. Consequently, this attribute leads intelligence to exhibit both open and closed characteristics that facilitate intelligent behavior \cite{shadab2024closed}. To support this perspective, we have introduced the "Core and Periphery" precepts, leveraging foundational principles from early cybernetics to enable systems modeling from both open and closed perspectives \cite{cody2022core}. 

This paper builds upon our earlier work by further exploring the role of core and periphery precepts in open-system and closed-system modeling. Through empirical examples, we illustrate the presence and relevance of these concepts across various intelligent system archetypes \cite{panchal2017experiments, szajnfarber2020call}. Before diving into these details, we provide a concise summary of the key cybernetic concepts that serve as the foundation for our work will be provided here.

Early cyberneticists, most notably Ashby, harnessed the concept of "variety" to model phenomena within complex systems. In his studies, Ashby specifically employed the concept of variety to explore homeostasis, which refers to a system's capacity to maintain specific variables within narrow bounds, even in the face of changing contexts, focusing on biological systems \cite{ashby1961introduction, ashby1968principles, klir1991requisite}.

Variety, as understood here and throughout, signifies the count of distinct elements. In this context, \emph{system} variety pertains to the number of unique elements within a system, while \emph{context} variety characterizes the number of unique elements present within the system's context \cite{schwaninger2023variety}. Typically, variety is defined concerning states, but it also extends to encompass inputs, denoted by $\mathcal{X}$, and outputs, denoted by $\mathcal{Y}$.

The main result of Ashby's research is a principle termed the \emph{Law of Requisite Variety} \cite{klir1991requisite}. It states that for a system to be stable, the system variety must be greater than or equal to the variety in the context it regulates. Formally put, given a system $S$ and context $C$,
\begin{equation}
\label{eqn:cond1}
     S \emph{ is stable} \to \emph{Variety}(S) \geq \emph{Variety}(C).
\end{equation}

Ashby presented a second, less-referenced, but similarly fundamental result. Let an outcome be an appearance of a particular element or co-occurrence of elements in the context. The variety of possible outcomes is lower-bounded by the difference between system variety and context variety. In other words, a system's (relative) variety limits the precision with which it can regulate phenomena in the context. Formally put, for a system $S$, context $C$, and outcomes $O$ \cite{ashby2011variety},
\begin{equation}
\label{eqn:cond2}
    \emph{Variety}(O) \geq \emph{Variety}(C) - \emph{Variety}(S).
\end{equation}
Naturally, $\emph{Variety}(O) \geq 0$, so in the case where system variety is greater than context variety the lower-bound is 0. 

In summary, Condition \ref{eqn:cond1} suggests stability requires a system's variety to be able to scale to match the variety of a system's contexts. Condition \ref{eqn:cond2} suggests that when the context variety is not well-matched by system variety, the set of possible outcomes is necessarily large and the system will struggle to achieve precise regulation of its context.

The Law of Requisite Variety found immediate and long-lasting use in organization management \cite{raadt1987ashby, ostrom1995designing, boisot2011complexity}. Recently, biological theoreticians have used requisite variety to posit how the mind emerges from the brain in a theory termed practopoiesis \cite{nikolic2017deep}, and have suggested that deep learning methods lack the variety required to scale to human-level intelligence \cite{nikolic2016testing}. Though not explicitly connected, the importance of variety is echoed in learning theoretic notions of capacity and local learning \cite{bottou1992local} as well as in emerging notions of free energy minimization and active inference \cite{friston2010free}.

We assert that the Law of Requisite Variety can be effectively harnessed to structure Systems Engineering (SE) practices, distinguishing between open and closed paradigms, particularly within the context of intelligent systems. To facilitate the practical application of this concept, we have introduced the "core and periphery" precepts. This framework enables the division of systems into core and periphery elements, allowing them to be methodically engineered in accordance with the principles of open-view and closed-view SE, as deemed appropriate. Subsequently, we present compelling evidence that underscores the relevance of the core and periphery precepts in modeling the behaviors and structures of intelligent systems.

The paper's structure is organized as follows: In the following section, we will provide a concise recap of our previous work concerning the formalization of the core and periphery precepts, elucidating its derivation within the context of the Law of Requisite Variety. Following that, we will offer a brief overview of how intelligent systems can benefit from the core and periphery framework, emphasizing the distinct characteristics of their structure, functionality, and behavior. Next, we reinforce our argument with real-world examples that illustrate how these systems can be effectively represented as core and periphery components. Finally, we provide a mathematical representation of core-dominant and periphery-dominant systems.

\section{Background}


In this section, we offer a comprehensive overview of the formal definitions of key concepts, as derived from the Law of Requisite Variety in our prior work, employing the perspective of general systems theory \cite{cody2022core}. The general systems theory's lens facilitates the application of the concept of variety in SE practices.

In the domain of general systems theory, systems are typically characterized as relations among sets. General systems theory predominantly deals with the fundamental principles governing relationships within sets. These principles can span across categories, topologies, algebraic structures, and more. However, it is important to note that the study of set theory, on its own, can provide valuable insights into the fundamental nature of these specific concerns. 
As mentioned earlier, Ashby introduced a particular concept of "variety" to investigate homeostasis in biological systems \cite{ashby1961introduction, ashby1991requisite}. In our previous work, we provided a set-theoretic redefinition of the concept of "Law of Requisite Variety".

Consider two systems $S$ and $S_E$ where $S:\mathcal{X} \to \mathcal{Y}$ and $S_E: \mathcal{X}_E \to \mathcal{Y}_E$. Without loss of generality term $S$ the system and $S_E$ the context. Suppose $S$ is acting as a regulator of $S_E$. Let $\mathcal{X}_{E \setminus S} = \mathcal{X}_E \setminus \mathcal{Y}$ where $\setminus$ denotes set difference. In other words, inputs to the context $\mathcal{X}_E = \mathcal{X}_{E \setminus S} \cup \mathcal{Y}$. Consider a set of outcomes $\mathcal{Z}$ with support over $\mathcal{X}_{E \setminus S} \times \mathcal{Y}$, i.e., $\mathcal{X}_{E \setminus S} \times \mathcal{Y} \to \mathcal{Z}$. 
Alternatively, variety can be expressed in terms of information complexity, a concept that Ashby quantified using the formula $log_2(n)$, where $n$ corresponds to the count of unique elements \cite{ashby1961introduction}. Let $V_A$ be termed \emph{variety} and be the Shannon entropy of a finite set $A$, i.e., 
\begin{equation}
\label{eq:var}
    V_A = - \sum^{|A|}_{i} p_i \log_2 p_i
\end{equation}

Where $|A|$ denotes the cardinality of $A$ and $p_i$ the probability of the $i^{th}$ element of $A$. Variety describes the number of unique elements in a system. The "Law of Requisite Variety" stipulates that in order for one system to effectively serve as a stable regulator for another, the variety present in the regulator's output must be either greater than or at least equal to the variety in the input of the system being regulated. In a more explicit formulation, this law posits that $V_Y$ (referring to the regulator's variety) must surpass or be on par with $V_{X_{E \setminus S}}$ (the variety associated with the context's input) for the achievement of precise, well-defined outcomes. This law implies that when the variety in the system's output fails to match with the variety present in the context's input, it results in a broader spectrum of possible outcomes, thereby making the attainment of precise results more challenging. In the words of Ashby, this law essentially conveys that a system's "capacity as a regulator cannot exceed its capacity as a channel for variety" \cite{ashby1991requisite}. Formally put, consider that (from \cite{ashby1991requisite})


\begin{equation}
    \min V_Z = \max \{ V_{\mathcal{X}_{E \setminus S}} -
    V_{\mathcal{Y}}, 0 \}.
    \label{eq:req}
\end{equation}

To help formalizing core and periphery concept, the Law of Requisite Variety can be redefined as follows.
\begin{definition}[Law of Requisite Variety]
The \emph{Law of Requisite Variety} states that given $V_{\mathcal{X}_{E \setminus S}}$, the minimum variety of outcomes $\min V_Z$ only decreases if $V_Y$ increases.
\end{definition}

Only if $V_{Y} \geq V_{X_{E \setminus S}}$, is it information theoretically possible to determine outcomes $\mathcal{Z}$, i.e., $\min V_Z = 0$.

Ashby linked a system's survival to the bounding of varieties \cite{ashby1961introduction}. Bounded varieties refer to system varieties that remain constant, while unbounded varieties pertain to those that change. This distinction allows us to identify core components of a system with bounded varieties, and the peripheral components with unbounded varieties. In the following, we will delve into how the Law of Requisite Variety serves as the foundation for establishing the core and periphery precepts.

To define core and periphery, let $S$ be a system $S \subset \times \{\mathcal{X}, \mathcal{Y}\}$ and let $\overline{S}$ denote the component sets of $S$, i.e., $\{\mathcal{X}, \mathcal{Y}\}$. Let $\mathcal{X}^t$ denote the input structure at time $t$, and so forth. Bounded and unbounded varieties are distinguished by measuring the variety of a system's residual change over time. Let $R$ denote this residual change, i.e.,
\begin{equation}
\label{eq:residual}
    R_{\overline{S}}^{t, t'} = \{\mathcal{X}^{t'} \setminus \mathcal{X}^t \,,\, \mathcal{Y}^{t'} \setminus \mathcal{Y}^t\}
\end{equation}
$R_{\overline{S}}^{t, t'}$ gives the residual change in system structure between time $t$ and $t'$. The core and periphery are defined as follows.
\begin{definition}[Core and Periphery]
Consider a system $S$ at time $t$ and at a later time $t'$. The \emph{core} of $S$ from $t$ to $t'$ is 
\begin{equation}
    \mathcal{C}_{\overline{S}}^{t, t'} = \overline{S} \setminus R_{\overline{S}}^{t, t'}
\end{equation}
The \emph{periphery} of $S$ from $t$ to $t'$ is 
\begin{equation}
    \mathcal{P}_{\overline{S}}^{t, t'} = R_{\overline{S}}^{t, t'}.
\end{equation}
\end{definition}

In simple terms, the core are those elements of $S$'s component sets that are identical at times $t$ and $t'$, and the periphery are those elements that are not. As a result, the core and periphery precepts can be utilized to model relative balance of variety in system and context as well as to understand how system addresses variety in the context \cite{cody2022core}.

In summary, the Law of Requisite Variety serves as a fundamental principle underpinning our core-periphery precepts. In the next section, we will leverage the formalization of these concepts to explore the unique features of intelligent systems that make them particularly amenable to core-periphery modeling. We will articulate how the core and periphery precepts are applicable in modeling the structure, behavior, and organization of intelligent complex systems. 

\section{Intelligent Systems}
The assessment of relative balance of variety in system and its context, as well as the system's adaptation to context variety, can be longitudinally observed to comprehend the learning or evolution of  the system-context relationship in terms of core and periphery. With this perspective, we propose that core and periphery constitute pertinent principles for the engineering of intelligent systems, particularly when intelligence is defined as an attribute intrinsic to the interaction between the system and its context. 

\subsection{Open and Closed System Phenomena}

\begin{figure*}[t]
    \centering
    \includegraphics[width=\textwidth]{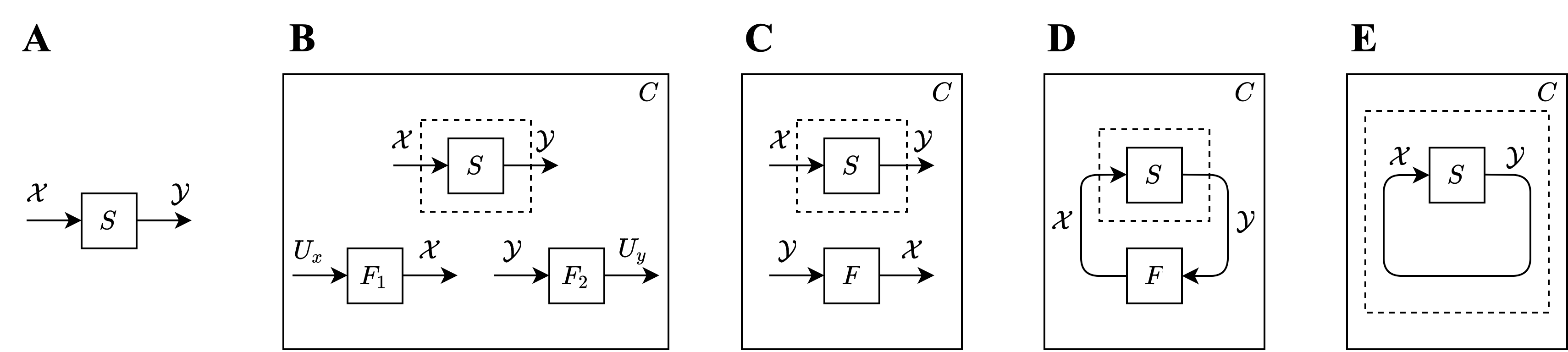}
    \caption{A depiction of the spectrum between open systems (A) and closed systems (E).}
    \label{fig:mesh3}
\end{figure*}

Open systems and closed systems precepts are important concepts in the SE practices \cite{wymore2018model}. Systems can be engineered with different spectrum of openness and closedness. On one extreme, as shown in \ref{fig:mesh3}-A, an open system $S$ exists in isolation, with no notion of context other than as everything outside the boundary. Loosening this, there exists a scoped context $C$ around $S$, and, further, there are other open systems in $C$ that receive $S$'s outputs $\mathcal{Y}$ or produce $S$'s inputs $\mathcal{X}$, as shown in Figure \ref{fig:mesh3}-B. The context is prepared for closure when $S$ is able to discover how other systems combine to produce $\mathcal{X}$ from $\mathcal{Y}$, as shown in Figure \ref{fig:mesh3}-C. $S$ can use this relation to achieve closure, as shown in Figure \ref{fig:mesh3}-D. The coupling becomes stronger and ultimately reaches the point where the system producing $S$'s inputs is integrated with $S$ itself. As shown in Figures \ref{fig:mesh3}-E, the scope of $S$ expands to the limits of $C$, and $S$ becomes a closed system, the other extreme.

Closed systems $S$ have inputs $\mathcal{X}$ that are influenced by their outputs $\mathcal{Y}$ by way of context $C$. Intelligent systems $S$ are expected to influence their context $C$. And since the closure of $S$ is mediated by $C$, $S$ is expected to influence the nature of their closure with $C$. In other words, intelligent systems have influence over their coupling with context. This suggests that even if intelligence can be relegated to a subsystem at conception, the boundaries between the intelligent system and those under its influence face dissolution as the intelligent system and its context intertwine. In closed view, in some ways, the system is just a vessel for intelligent subsystems to gain influence over the context. 



\noindent The open-system view may suffice in certain contexts, but as algorithms and their applications become increasingly intricate, the closed-system view might become indispensable as the closed effects of the context and system will rise.

Putting open-view and closed-view dichotomy aside, consider now the distribution of open- and closed-system phenomena across the core and periphery. The core is given by system varieties that are invariant to context. So, the core is decoupled from context. It follows then that the core corresponds to open system phenomena. By the same logic, the periphery corresponds to the part of the system coupled with context, i.e., to closed system phenomena. Importantly, while a distributed core may be formed by boundary-dissolving closed system phenomena, once it forms, its invariance to changes in context characterize it by open system phenomena. 

Adaptation to changes in context involves interplay between the core and periphery precepts, and thus interplay between open-view and closed-view modeling of the system. To see this, consider that the periphery, by way of its coupling, is the part of the intelligent system that initially responds to changes in context. Put differently, by definition, the core is invariant to changes in context, so new information must enter by way of the periphery. This leads to the first principle.

\begin{principle}
    The adaptation of intelligent systems to changes in context is principally a closed system phenomena (from the periphery) resulting in open system phenomena (in the core).
\end{principle}
\noindent Principle 1 suggests that in order to adapt the core to new information, new varieties from the periphery must be identified and progressively bounded in a process that involves both closed and open system phenomena.

\subsection{Scaling Intelligence}

The core and periphery precepts could be utilized to explain the scale of intelligent property in systems. Addressing the scalability of intelligence is perhaps the strongest motivation for orienting the engineering of intelligent systems towards the concepts of core and periphery. Presently, the notion of scale in Artificial Intelligence (AI) and Machine Learning (ML) systems primarily centers around parameters like the number of queries per second, the number of user interactions per second, or data processing volume \cite{haefner2023implementing}. In essence, scalability is often measured by a system's ability to handle increased input and output complexity \cite{bano2023artificial}. However, this perspective on scaling is rather limited.

A more comprehensive approach to measuring intelligence's scalability needs to consider the system's variety. Expanding the number of input and output elements results in the scaling of variety. Nonetheless, it's crucial to recognize that scaling inputs and outputs represents just one special case of scaling system variety, and it is not the most generalized one. Focusing solely on scaling inputs and outputs aligns with an open-system perspective.  A focus on inputs and outputs corresponds to scaling bounded variety, but, at some point, scaling intelligence requires engineering intelligent systems described predominantly by unbounded variety \cite{shadab2022closed}. In other words, when confronted with the challenge of scaling intelligence, engineering intelligent systems that primarily adhere to closed system phenomena becomes essential \cite{shadab2022closed}. 

To illustrate this, imagine a scenario where the context's variety grows indefinitely with increased scaling. Eventually, by logical necessity, the system's variety, which matches its bounded varieties with unbounded ones in the context, cannot maintain stability or attain precise outcomes. Consequently, the onus of scaling intelligence gravitates toward the system's unbounded varieties – its periphery – and, in turn, toward closed system phenomena. This insight leads to the second fundamental principle:

\begin{principle}
    At some point, scaling the variety of an intelligent system relies predominantly on closed system phenomena.
\end{principle}

Yet, one might wonder why it is crucial to scale variety. Equations \ref{eqn:cond1} and \ref{eqn:cond2} make it evident that scalability is a prerequisite for both stability and achieving precise outcomes. 
The strong couplings between systems and their contexts essentially prevent us from establishing a clear boundary between system inputs and outputs when the coupling is potent. The concept of closure can be used in modeling process in the extreme case of strong coupling, where the system's boundary dissolves with the context \cite{shadab2024systems}. This emphasizes the importance of adopting precepts that aren't contingent on steadfast system boundaries \cite{shadab2024systems}. Therefore, given the close relationship between coupling and scale, the core and periphery concepts prove their relevance as essential precepts for engineers when scalability in intelligence is the primary objective.

\section{Empirical Evidence of Core-Periphery Precepts in Real-World}

In this section, we provide some examples for the core and periphery precepts manifested in real-world systems. Our goal is to unveil and meticulously examine the persuasive evidence supporting the feasibility of modeling structural, behavioral, or organizational aspects within these intricate real-world systems through the lens of core-periphery concepts.

\subsection{DNA and Neurons}
The DNA structure encapsulates key features of an intelligent system; however, its slow adaptation to changes unfolds over generations, influencing the fundamental characteristics of the intelligent system \cite{wagner2013robustness}. The alterations in DNA structure, essential for defining the system's traits, undergo a gradual process that spans multiple generations. In contrast, the intricate coordination and communication among neurons in the brain enable a broad spectrum of tasks, ranging from simple reflexes to complex cognitive functions like problem-solving and decision-making \cite{zhang2019cognitive}.

Neurons serve as both information processors and signal transmitters, facilitating the brain and nervous system's effective response to the environment and the execution of various tasks. This structural arrangement allows intelligent systems to maintain core characteristics, such as the number of legs or skull structure, while also providing flexibility for evolution in response to environmental demands. The brain's ability to increase neuronal connections contributes to the system's variety, enabling adaptations to the environment. Over time, these adaptive changes, driven by the brain, influence the DNA to transition towards a more stable core by addressing emerging needs arising from contextual adaptation \cite{schwaninger2023variety}.

The concept of "survival of the fittest" is rooted in the idea that genetic variations leading to better adaptation to the environment increase the likelihood of survival and subsequent inheritance in subsequent generations \cite{paul1988selection}.

The brain's capacity for self-reorganization, characterized by the formation of new neural connections, is an example of periphery. This capability enables adaptation through learning, recovery from injury, and adjustments to changes in sensory input. In contrast, DNA represents a tightly bounded variety. The nervous system, acting as the periphery, exhibits a lesser degree of constraint. This conceptualization gives rise to the core and periphery framework in intelligent systems—featuring a core set of tightly bounded varieties enveloped by layers of progressively less constrained varieties extending into the system's periphery, where variety is unbounded. The core and periphery dynamics, in terms of survival, involve preventing disorder from penetrating the periphery deeply enough to reach the core. The approach to model the core and periphery diverges significantly based on whether one adopts an open or closed view. 

\subsection{Homeostasis vs Homeodynamic}
In the study of biological systems, two fundamental concepts are often employed to describe stability and equilibrium: homeostasis and homeodynamics \cite{yates1994order, cannon1932homeostasis}. These concepts are essential for understanding how living organisms maintain stability in different ways.

Homeostasis is a fundamental physiological and biological concept that refers to the body's ability to maintain a stable and balanced internal environment, despite external changes and fluctuations \cite{reimann1996homeostasis}. It involves the regulation of various factors such as temperature, pH levels, blood pressure, and the concentration of nutrients and gases within the body \cite{torday2015homeostasis}. The key principle of homeostasis is to keep these internal conditions within a narrow and optimal range, which is essential for the proper functioning of cells, tissues, and organs \cite{cannon1932homeostasis}. When external factors or internal processes disrupt this balance, the body employs various mechanisms to restore equilibrium. For example, if body temperature rises due to external heat, mechanisms like sweating and dilation of blood vessels help cool the body down. Conversely, if body temperature drops in a cold environment, shivering and constriction of blood vessels help generate and conserve heat.

On the other hand, homeodynamic is a term used to describe a state of dynamic equilibrium or balance within a biological system \cite{lloyd2001homeodynamics}. Homeodynamic points out to the biological systems' ability to dynamically self-organise at bifurcation points of their behaviour where they lose stability \cite{yates1994order}. For example, the immune system is highly dynamic. It can mount a rapid response to pathogens, adapt to new threats, and return to a resting state once the threat is eliminated. This dynamic equilibrium allows the immune system to protect the body without causing excessive inflammation. Another example of dynamic equilibrium is metabolism. Metabolism rate can have dynamic equilibrium that actively adjust with the body needs and activities.

Homeostasis can be viewed as the result of spatio-temporal chaotic dynamics, as discussed by Ikegami \cite{ikegami2008homeostatic}. In this context, the homeodynamic variables within biological systems exhibit temporal changes, while the parameters governing homeostasis remain constant over time \cite{ikegami2008homeostatic}. Drawing a parallel to the core and periphery framework, we can equate homeostasis process aligned with the core, while homeodynamics process finds its place in the the periphery model within these biological systems. This alignment strongly indicates the presence of a specific organizational structure within biological systems that lends itself to modeling through the core and periphery principles.






\subsection{Interrogative Attitude vs Belief} 

This paper delves into a philosophical discourse concerning the nature of inquiry and its role on transforming interrogative attitudes into beliefs and vice versa. Within this philosophical context, we elucidate how the cognitive dynamics of an intelligent mind operate and, notably, how these facets correlate with the previously introduced conceptual framework of core and periphery within this paper.


Suspension of judgment; a state of mind that one withholds judgement \cite{friedman2013suspended}; emerges as an integral facet closely intertwined with the process of inquiry \cite{friedman2019inquiry}. It is imperative to recognize that the act of suspending judgment is a deliberate step taken in the pursuit of genuine inquiry. Inquiry, in this context, is defined as an interrogative attitude, representing a specific cognitive state of mind. Within the realm of interrogative attitudes lie various psychological states and processes, encompassing curiosity, wonder, contemplation, deliberation, and more \cite{friedman2019inquiry}. These attitudes can manifest as either static states or dynamic processes. Consequently, inquiry emerges as a manifestation of a goal-oriented cognitive state that necessitates corresponding actions to foster and conduct the inquiry. The essence of inquiry lies in the act of asking questions, which serves as a goal-directed endeavor aimed at acquiring specific epistemic information.

A crucial distinction emerges between belief and interrogative attitude. Belief signifies the possession of a complete answer to a given question, denoted as $Q_1$, at a specific time, $t_1$ \cite{friedman2019inquiry}. In contrast, an interrogative attitude represents a distinct cognitive state of mind wherein one actively seeks a comprehensive answer to a different question, labeled as $Q_2$, at a subsequent time, $t_2$. Notably, a belief held at time $t_1$ can transition into an interrogative attitude at a subsequent time, $t_2$, when a suspension of judgment occurs and additional information becomes available \cite{friedman2019inquiry}. Conversely, the initiation of a suspension of judgment marks the point at which the transformation begins. For instance, a detective aiming to identify the criminal in a case often starts with an initial suspect and poses goal-oriented questions while exploring additional evidence. Throughout this process, the detective may exhibit various attitudes such as questioning, curiosity, and wonder. However, as the investigation progresses and sufficient inquiries are addressed, a definitive belief about the guilt of a particular individual can be established.

In this context, it is essential to recognize that interrogative attitudes undergo more frequent fluctuations, and these shifts in attitude through inquiries inherently culminate in a transformative process that ultimately influences changes in one's beliefs. Beliefs, conversely, tend to exhibit a higher degree of stability \cite{hoek2020minimal}. This stability persists unless the ongoing inquiry successfully yields a comprehensive answer to the posed question.

The distinction between interrogative attitude and belief further underscores their fundamental disparities. Interrogative attitude, by nature, adheres to a closely defined and goal-oriented paradigm. It operates as an active and dynamic cognitive endeavor, inherently focused on a specific objective. In contrast, belief adopts a more open-view conceptual framework. Belief operates akin to a function, mandating the attainment of a complete answer, represented as outputs, to a given question, denoted as inputs \cite{leitgeb2014stability}. This distinction underscores the fundamental dichotomy between interrogative attitude and belief.

Essentially, the interrelationship between interrogative attitude and belief encapsulates a dynamic interplay, wherein the evolving attitude often acts as the harbinger of alterations within one's belief system. This dynamic can be aptly modeled or conceptualized through the framework of the core and periphery precepts. Within this context, beliefs are a set of states that can be modeled as bounded varieties encompassed by the core. This core represents the central and more stable component of the belief system. Conversely, interrogative attitudes assume the role of the periphery of the belief system. Periphery represents the dynamic and ever-changing aspect of one's cognitive framework within this system.


\section{Evidence of Core-Periphery Precepts in Engineered Intelligent Systems}

In the previous section, we presented supporting evidence for the applicability of the core and periphery precepts in modeling biological systems. In this section, we delve into analogous evidence within engineered intelligent systems.

We assert that the structure of even the simplest intelligent systems exhibits evidence of the applicability of the core and periphery precepts. To substantiate this claim, we devised an experiment involving a Convolutional Neural Network (CNN). The aim was to assess alterations in the weights of its fully connected layer when confronted with significant changes in context through its inputs set.

\begin{figure}[h]
    \centering
    \includegraphics[scale=0.29]{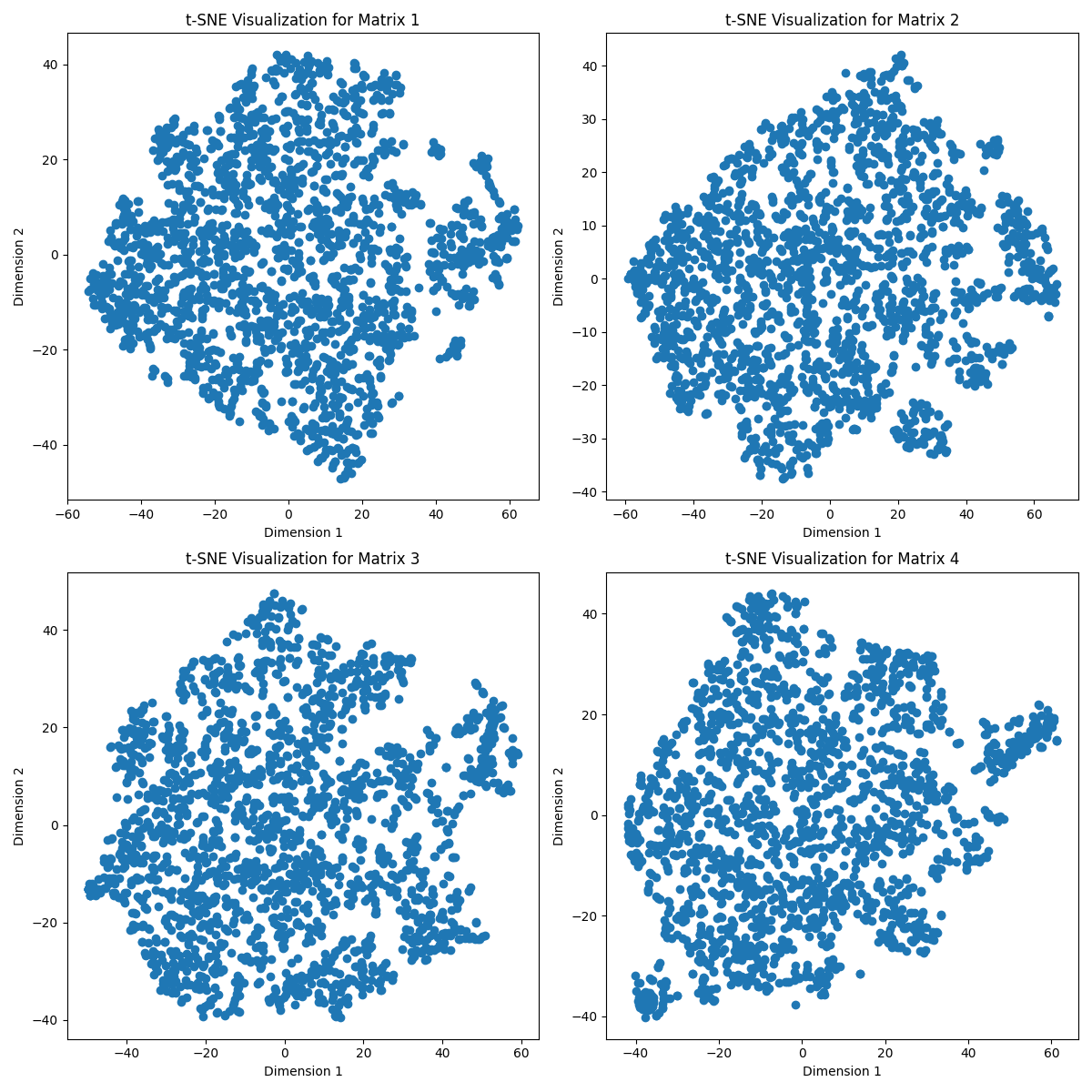}
    \caption{The Comparison of the weights difference of $1^{st}, 10^{th}, 20^{th}, 30^{th}$ with the weight difference of $40^{th}$ epoch of the second model. The weights of the epochs in the second model were already subtracted from the $40^{th}$ epoch of the first model}
    \label{separated_t-sne}
\end{figure}

The experiment was structured as follows: Utilizing the ResNet-50 model, an existing CNN model from Tensorflow \cite{koonce2021resnet}, we conducted training on the CIFAR-10 dataset. This dataset encompasses 60,000 images of size 32$\times$32, distributed across 10 classes. Throughout the training of the ResNet-50 model, we recorded the weights of the fully connected layer for each epoch, spanning a total of 40 epochs.

\begin{figure}[h]
    \centering
    \includegraphics[scale=0.63]{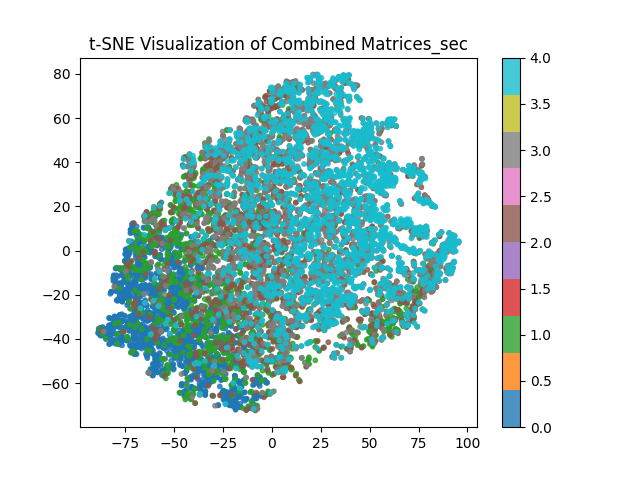}
    \caption{The t-SNE visualization of combined weights of $1^{st}$ epoch, $10^{th}$ epoch, $20^{th}$ epoch, $30^{th}$ epoch, $40^{th}$ epoch from the second train of the ResNet-50 after we subtracted the weights of the $40^{th}$ epoch of the first run with each of the epochs mentioned above}
    \label{t-sne-combined}
\end{figure}


Subsequently, we employed the previously trained model, with its stored weights, and retrained it using CIFAR-100, a dataset of similar nature but with increased complexity and a total of 100 classes. This modification in the model reflects potential alterations in the context that can be simulated by changes in the input set of a CNN model. We recorded the weights of the fully connected layer for each epoch during the retraining of the model, extending over a total of 40 epochs. This experiment enables a comparative analysis of the weights across epochs of the second trained model with those of the first trained model, shedding light on any discernible patterns in the changing weights of the fully connected layer. 

In this experimental setup, weights serve as a simplified representation of system variety. Layers with minimal to no changes in weights can be conceptualized based on the principles of the core precept (open-systems view). Conversely, layers exhibiting significant weight variations in response to changes in the input set (context variety) can be conceptualized through the periphery precept. It is essential to emphasize that this example serves as an illustration of such conceptualization within a CNN model. The primary objective is not to assess the practical utility of modeling the fully-connected layer with core and periphery precepts. As previously mentioned, the core and periphery model is most fitting for types of intelligence that represent a highly coupled relational property between the system and its context.

\begin{figure*}
    \centering
    \begin{subfigure}[h]{0.5\textwidth}
        \centering
        \includegraphics[height=2.9 in]{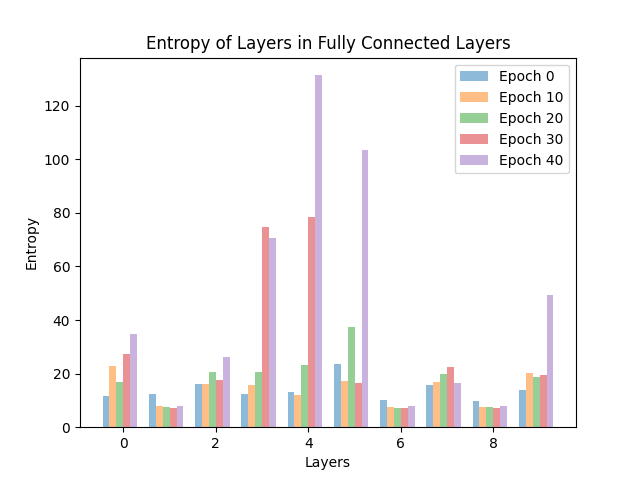}
    \end{subfigure}%
    ~ 
    \begin{subfigure}[h]{0.5\textwidth}
        \centering
        \includegraphics[height=2.9 in]{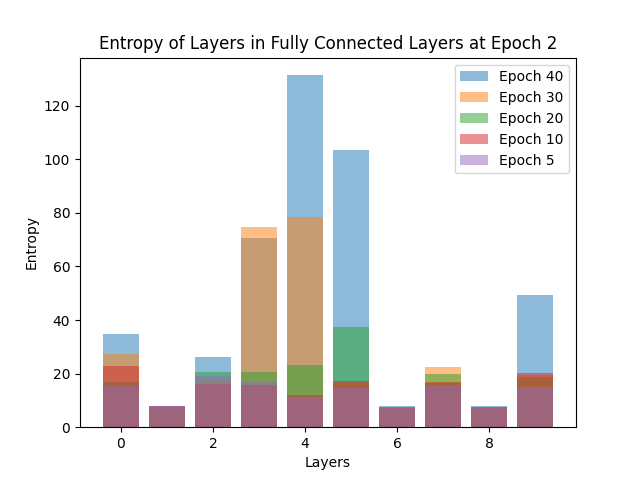}
    \end{subfigure}
\caption{The comparison between Entropy estimation of each Layer in Fully Connected Layers, as epochs progress. For demonstration purposes only, Epochs 0, 5, 10, 20, 30, 40 are captured in the plot}
\label{entrop-combined}
\end{figure*}

Figure \ref{t-sne-combined} presents a visualization of the combined weight distribution of the epochs in the second model. We utilized t-SNE, a dimension reduction technique, to facilitate the visualization of the high-dimensional data from the fully connected layers of ResNet-50. Initially, we selected four epochs from the second training model and computed the weight differences by subtracting the weights of these epochs from the $40^{th}$ epoch of the first model. Subsequently, we compared these weight differences using t-SNE visualization. As depicted in Figure \ref{t-sne-combined}, the weights shifted from negative to positive dimensions as the epochs progressed. However, a central region with minimal dimension differences between all epochs is observable. For improved visibility, we isolated the weight differences of the $1^{st}, 10^{th}, 20^{th}, 30^{th}$ epochs and compared them with that of the $40^{th}$ epoch from the second model. Figure \ref{separated_t-sne} illustrates that the first and second comparisons (top left and top right) exhibit weights shifted to the left, while in the third and fourth comparisons, the weights shifted to the right. Nevertheless, consistent patterns that remained unchanged across all four comparisons are discernible.

To examine our core and periphery concept and illustrate the observed patterns in the fully-connected layers of the model, we aimed to quantify the weight changes at each individual layer as the epochs progressed. This approach allows us to argue that layers with smaller changes in neuron weights represent the core layers, while those with larger rates of change between epochs form the periphery of the structure. As discussed in Section II, core and periphery are defined through Variety, which can be quantified using the Shannon entropy formula. Thus, the next step in our simulated experiment was to calculate and track the changes in entropy at each layer to determine Variety, and, consequently, the core and periphery structure.

To calculate Shannon entropy, we constructed a weight matrix for each layer. We then normalized the singular values of each weight based on the sum of all weights in that layer, denoted as $p_i$, to form a probability distribution. These probabilities ($p_i$) were then used in Equation \ref{eq:var} to estimate entropy based on the changes in weight distribution for each layer at each epoch. Figure \ref{entrop-combined} illustrates the entropy changes across all layers in the fully-connected structure.

As shown in Figure \ref{entrop-combined}, at least three layers exhibit minimal entropy change as the epochs progress, while the middle layers experience the most significant spike in entropy with increasing epochs. It is important to note that, when calculating entropy at Epoch 0, the weights were subtracted from those of the initial model. This approach indicates that by starting with a pre-trained model and retraining it with a new dataset, the existing entropy from the original model is carried over to the retrained model. Consequently, the entropy at the first epoch already considers and factors in the existing entropy in the model. Thus, all entropy changes depicted in Figure \ref{entrop-combined} capture the core and periphery structure of the model after the introduction of a new dataset. As previously mentioned in this section, a new dataset represents disturbances or changes in the context of the engineered intelligence system.

From this analysis, we conclude that even in the simplest forms of engineered intelligent models, patterns of core and periphery are still manifested. This finding suggests the potential to engineer desired intelligent system outcomes by strategically designing the core and periphery structures.





\section{Core-Dominant vs Periphery-Dominant Systems}

To bound the outcomes of an intelligent system, it is consequential to determine if the system is being modeled as a core-dominant system or periphery-dominant one. Core-dominant model of the system indicates that the characteristics of the system depends more on the behaviors (outputs) of the core whereas periphery-dominant model of the system relies mostly on the behaviors (outputs) of the periphery part. In this section, we will elucidate the conditions for having a core-dominant vs a periphery-dominant model of a system.

\subsection{Core-Dominant Systems}

Core-dominant systems are characterized by having greater variety in their core compared to their periphery, denoted as $V_{\mathcal{C}} \geq V_{\mathcal{P}}$. This classification is attributed to the core's role in impeding more variety from the system's environment \cite{cody2022core}. As mentioned earlier, Shannon's entropy can be used to capture variety following Equation \ref{eq:var}. The entropy of a system will be determined by the joint entropy of the core and periphery parts of the system. From the entropy equation $H(X,Y) = H(X) + H (Y|X)$ \cite{cover1991entropy}; we can derive:

\begin{equation}
\begin{split}
    H(\overline{S}) = H(\mathcal{C}_{\overline{S}}^{t, t'}) + H(\mathcal{P}_{\overline{S}}^{t, t'} | \mathcal{C}_{\overline{S}}^{t, t'}) \\
    = H(\mathcal{P}_{\overline{S}}^{t, t'}) + H(\mathcal{C}_{\overline{S}}^{t, t'}|\mathcal{P}_{\overline{S}}^{t, t'})
    \label{ashby}
\end{split}
\end{equation}

As mentioned earlier, core is not prone to residual changes of inputs and outputs. Without loss of generality, if $\mathcal{C}_{\overline{S}}^{t, t'}$ is the core of our system; $\overline{S}$; we can have a core-dominant model of the system if we have this constraint as follows:

\begin{equation}
    H(\mathcal{C}_{\overline{S}}^{t, t'}|\mathcal{P}_{\overline{S}}^{t, t'}) \leq H(\mathcal{C}_{\overline{S}}^{t, t'}) \leq \delta
\label{core-dominant}
\end{equation}

$\delta$ represents a modeling threshold, specifying that the conditional entropy of the core given the periphery should be less than or equal to a certain value. This condition ensures that there is adequate entropy (i.e., variety) in the core to exhibit core-dominant characteristics in the system's model while maintaining the entropy of $\overline{S}$ at an acceptable level. Equation \ref{core-dominant} is derived from the relation between the system ($\overline{S}$) and its consisting parts ($\mathcal{C}_{\overline{S}}^{t, t'}$, $\mathcal{P}_{\overline{S}}^{t, t'}$). 

Equation \ref{core-dominant} is a result of simple assumptions and information theoretical equations as follows:
    \begin{align*}
        &\text{Per Core-dominant Definitions:}\\
        &H(\mathcal{C}_{\overline{S}}^{t, t'}) \geq H(\mathcal{P}_{\overline{S}}^{t, t'}) \\
        &\text{From Equation \ref{ashby} and inequality above: }\\
        &H(\mathcal{C}_{\overline{S}}^{t, t'}|\mathcal{P}_{\overline{S}}^{t, t'}) \geq H(\mathcal{P}_{\overline{S}}^{t, t'}|\mathcal{C}_{\overline{S}}^{t, t'})\\
        &\text{From information theory and inequality above:}\\
        &H(\mathcal{P}_{\overline{S}}^{t, t'}|\mathcal{C}_{\overline{S}}^{t, t'}) \leq H(\mathcal{P}_{\overline{S}}^{t, t'}) \leq H(\mathcal{C}_{\overline{S}}^{t, t'}|\mathcal{P}_{\overline{S}}^{t, t'}) \leq H(\mathcal{C}_{\overline{S}}^{t, t'})
    \end{align*}

According to the definition of core-dominant system provided in Equation \ref{core-dominant}, the system's characteristics is more dependant on the the core characteristics. This interpretation infers that not only the core has more entropy (i.e., variety) but also the conditional entropy of the core given periphery; $H(\mathcal{C}_{\overline{S}}^{t, t'}|\mathcal{P}_{\overline{S}}^{t, t'})$; will be greater than the entropy of the periphery part. Therefore, core-dependant systems can be modeled predominantly with the open-view paradigm that depends on the input-output relations between the system and its environment.


\subsection{Periphery-Dominant Systems}

In contrast to core, periphery is prone to residual changes in inputs and outputs sets. Periphery-dependent systems by definition allow for these residual changes and grow and/or modify the possible sets of outcomes based on such changes in periphery. In a periphery-dominant model of systems, changes in the outcomes of systems depend more on the changes in the context than on the bounded structure and behavior of the core. Without loss of generality, consider $\mathcal{C}_{\overline{S}}^{t, t'}$; as the core of the system and the rest of the system will be the periphery part; $\mathcal{P}_{\overline{S}}^{t, t'}$. We can have a periphery-dominant model of the system with a condition as follows.

\begin{equation}
    H(\mathcal{P}_{\overline{S}}^{t, t'}|\mathcal{C}_{\overline{S}}^{t, t'}) \leq H(\mathcal{P}_{\overline{S}}^{t, t'}) \leq \gamma
    \label{periphery}
\end{equation}

$\gamma$ represents a modeling threshold, specifying that the conditional entropy of the periphery given core should be less than or equal to a certain value. This condition ensures that there is adequate entropy (i.e., variety) in the periphery to exhibit periphery-dominant characteristics in the system's model while maintaining the entropy of $\overline{S}$ at an acceptable level.

Similar to Equation \ref{core-dominant}, Equation \ref{periphery} is a result of simple assumptions and information theoretical equations as follows:
    \begin{align*}
        &\text{Per Periphery-dominant Definitions:}\\
        &H(\mathcal{P}_{\overline{S}}^{t, t'}) \geq H(\mathcal{C}_{\overline{S}}^{t, t'}) \\
        &\text{From Equation \ref{ashby} and inequality above: }\\
        &H(\mathcal{P}_{\overline{S}}^{t, t'}|\mathcal{C}_{\overline{S}}^{t, t'}) \geq H(\mathcal{C}_{\overline{S}}^{t, t'}|\mathcal{P}_{\overline{S}}^{t, t'})\\
        &\text{From information theory and inequality above:}\\
        &H(\mathcal{C}_{\overline{S}}^{t, t'}|\mathcal{P}_{\overline{S}}^{t, t'}) \leq H(\mathcal{C}_{\overline{S}}^{t, t'}) \leq H(\mathcal{P}_{\overline{S}}^{t, t'}|\mathcal{C}_{\overline{S}}^{t, t'}) \leq H(\mathcal{P}_{\overline{S}}^{t, t'})
    \end{align*}

According to the definition of periphery-dominant systems provided in Equation \ref{periphery}, the system's characteristics is more dependant on the the periphery characteristics. This interpretation infers that not only the periphery has more entropy but also the conditional entropy of periphery given the core; $H(\mathcal{P}_{\overline{S}}^{t, t'}|\mathcal{C}_{\overline{S}}^{t, t'})$; is greater than the entropy of the core.

In conclusion, as stated in \cite{cody2022core}, without the need to have assumptions of functional dependence of the components of a system, as commonly done in traditional engineering practices that rely on decomposition and recomposition precepts, the principles of core and periphery can be employed. These principles offer a model that elucidates the aspects of an intelligent system used to constrain environmental variety and regulate outcomes.

\section{Discussion}


By definition, all systems—excluding perhaps the entirety of the universe—are inherently open systems that can be decomposed into smaller constituent subsystems \cite{parnas1972criteria, simon2012architecture}. These subsystems accept inputs in the form of energy, material, and information, and produce a set of predefined outputs \cite{maier_art_2009, crawley2015system}. To date, this fundamental characteristic has influenced the way engineered systems are modeled and realized using open systems principles. However, as demonstrated in this study through a combination of deductive and inductive evidence, core and periphery properties are ubiquitous across various classes of intelligent systems. These properties necessitate the consideration of both open and closed system principles in modeling such systems \cite{shadab2024systems}. Consequently, there is an urgent need for engineering practices to reevaluate decomposition mechanisms, particularly in terms of determining \textit{when} and \textit{how} system elements should be modeled based on the notions of \textit{openness} or \textit{closedness} to their behavior and/or structure. The core and periphery treats in the intelligent systems, allow engineers to model different properties based on the spectrum of their openness. This is also important to note that such spectrum can only be interpreted and evaluated through the systems' relations to their context. Belief vs interrogative attitude can only be modeled only if they are evaluated in a context. This paper discusses the potential implications of these findings for systems engineering practices and theory.

First, it is about articulation of system architectures, subsystem specifications, and contracting. SE practices are based on decomposition of system-level functions and their allocation to subsystems which are later contracted out or developed in house. Since complex systems are only partially decomposable \cite{simon2012architecture, simon2019sciences}, this process involves creation of interfaces\cite{topcu2022dark}, which help set the boundary and coordination rules across resulting modules. Creation and coordination of design work across these interfaces are documented to be leading source of design changes and cost growth, even in unintelligent systems \cite{blyler_interface_2004, sundgren_introducing_1999}. To that end, the presence of core and periphery properties, and particularly the scaling of dominance concepts discussed in Section VI, provide new opportunities for specification of module characteristics and interface rules in intelligent systems. By arranging these characteristics and defining to which extent a module should be modeled as closed or open, SE can guide how collaborating modules should be developed and how inputs and outputs should be shared across the boundaries. Although we did not attempt to take stab at how this can be operationalized, this notion could help guide the development effort by "locking in" some design features earlier in the process to regulate certain intelligent system behaviors over time \cite{doyle2011architecture}. 


Finally, core and periphery precepts could inform how intelligent systems can be developed to be more \textit{resilient} \cite{leveson2017engineering,uday2015designing}. Although definitions vary \cite{hosseini2016review}, engineering community broadly considers resilience as the ability of a system to endure and recover from external disruptions, through some form of an absorption, adaptation, and restoration of critical system functions over-time \cite{henry2012generic}. Therefore, for a system to be considered as intelligent, it should inherently exhibit resilience by learning from operational disruptions and changing its behavior accordingly. We contend that core and periphery properties, and allocation of system elements to these roles could help bolster a system's resilience, by categorizing which functions should be more adaptable to these events, and which should remain constant in its information change. The hope here is by considering these as design choices ahead of time, or operational interventions after deployment, SE could help guide intelligent systems to adapt as desired by its stakeholders, and not as a random mutation event that could be dictated by the systems' operational nature. 

\section{Conclusion}
In this paper, we have presented real-world examples demonstrating the practicality of the core and periphery precepts in various classes of intelligent systems. In our earlier work, we introduced and formalized these precepts, suggesting their potential utility in engineering intelligent systems through the application of open systems and closed systems principles. In this context, we conducted a small experiment aimed at unveiling the structure of a rather simple engineered intelligent system viewed through the core and periphery framework. These examples support our earlier deductive work by providing empirical evidence in a totally different set of intelligent systems, and indicate that not only can the behaviors and structures of biological systems be elucidated through the core and periphery precepts, but engineered intelligent systems also exhibit analogous structures and behaviors. We also introduced the concepts of core-dominant and periphery-dominant systems that can be utilized in modeling of systems with different emphasis on open-systems or closed-systems engineering principles.




\bibliographystyle{IEEEtranIES}
\bibliography{main}

\end{document}